\documentclass{article}

\usepackage{arxiv}

\usepackage[utf8]{inputenc} 
\usepackage[T1]{fontenc}    
\usepackage{hyperref}       
\usepackage{url}            
\usepackage{booktabs}       
\usepackage{amsfonts}       
\usepackage{nicefrac}       
\usepackage{microtype}      
\usepackage{lipsum}		
\usepackage{graphicx}
\usepackage{amsmath}
\usepackage{amssymb}
\usepackage{bm}
\usepackage{mathtools}
\usepackage[utf8]{inputenc}
\usepackage{doi}

\title{A Novel Edge Detection Operator for Identifying Buildings in Augmented Reality Applications}

\author{ \href{https://orcid.org/0000-0002-0071-958X}{\includegraphics[scale=0.06]{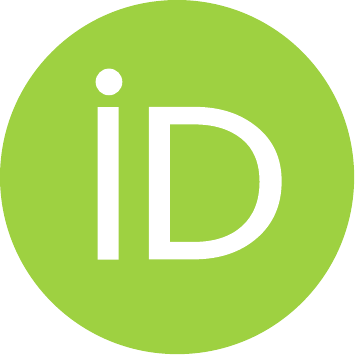}\hspace{1mm}Ciprian~Orhei}\\
	Politehnica University of Timi\c{s}oara\\
	Timi\c{s}oara, Romania\\
	\texttt{ciprian.orhei@cm.upt.ro} \\
	\And
	\href{https://orcid.org/0000-0003-2394-4859}{\includegraphics[scale=0.06]{orcid.pdf}\hspace{1mm}Silviu~Vert} \\
	Politehnica University of Timi\c{s}oara\\
	Timi\c{s}oara, Romania\\
	\texttt{silviu.vert@upt.ro} \\
	\And
	\href{https://orcid.org/0000-0003-1185-1997}{\includegraphics[scale=0.06]{orcid.pdf}\hspace{1mm}Radu~Vasiu} \\
	Politehnica University of Timi\c{s}oara\\
	Timi\c{s}oara, Romania\\
	\texttt{radu.vasiu@upt.ro} \\
}

\renewcommand{\shorttitle}{Accepted in 26th International Conference, ICIST 2020}

\hypersetup{
pdftitle={A Novel Edge Detection Operator for Identifying Buildings in Augmented Reality Applications},
pdfsubject={cs.CV},
pdfauthor={Ciprian~Orhei, Silviu~Vert, Radu~Vasiu},
pdfkeywords={augmented reality, building detection, edge detection, edge detection operator, Canny},
}

\begin{document}
\maketitle

\begin{abstract}
Augmented Reality is an environment-enhancing technology, widely applied in many domains, such as tourism and culture. One of the major challenges in this field is precise detection and extraction of building information through Computer Vision techniques. Edge detection is one of the building blocks operations for many feature extraction solutions in Computer Vision. AR systems use edge detection for building extraction or for extraction of facade details from buildings. In this paper, we propose a novel filter operator for edge detection that aims to extract building contours or facade features better. The proposed filter gives more weight for finding vertical and horizontal edges that is an important feature for our aim. 
\end{abstract}

\keywords{Building dataset  \and facade detection \and edge detection \and semantic segmentation \and edge detection ground-truth \and semantic segmentation ground-truth}

\section{Introduction}

Augmented reality (AR) is the process of overlapping computer-generated objects and data over our real, surrounding space \cite{milgram1994taxonomy}. This differs from virtual reality, where the basic elements of the environment are entirely computer-generated in an effort to simulate their existence \cite{wright2014using}. Augmented reality is successfully exploited in many domains nowadays, one of the them being culture and tourism, an area in which authors of this paper carried multiple research projects \cite{vert2014lod4ar, vert2014relevant, vert2017augmented, vert2019augmented}.

One of the main ways to detect the surrounding space in Augmented Reality is through Computer Vision (CV). This field of research has been developing mathematical techniques for recovery of the 3D shapes and appearances of objects in pictures. In particular, building extraction has been an active research topic in CV as well as digital photogrammetry. Building detection is the process of obtaining the approximate position and shape of a building, while building extraction can be defined as the problem of precisely determining the building outlines, which is one of the critical problems in digital photogrammetry \cite{elshehaby2009new}. Building information is extremely important not only for augmented reality applications, but also for urban planning, telecommunication, three-dimensional city modeling, or extraction of unauthorized buildings over agricultural lands \cite{elshehaby2009new}.

In this paper we present a novel filter operator for edge detection, section 3, that can be combined with an algorithm for contour detection that tries to enhance the topic above mentioned, building extraction. This is the first step towards efficient building detection in augmented reality, a research endeavor that we are currently undertaking as part of the Spotlight Heritage, a cultural project for Timisoara European Capital of Culture 2021 \cite{spt}.

In section 4 we make a visual comparison with other standard filters similar to the ones we propose: Sobel filter \cite{sobel19683x3}, Prewitt filter \cite{prewitt1970object}, Scharr filter \cite{scharr2000optimal}. Because we are looking for a particular use case, like buildings, we want to look at how the 5x5 extension of the mentioned filters perform. To better understand the capabilities of the filters we will use the Canny algorithm \cite{canny1986computational} which is one of the most popular algorithms for edge detection.

\section{Related work}
\label{Sec:related_works}

In the research literature we can see that applications concerning augmented reality in urban scenes have different approaches. In this section we will present different solutions that we came across that use edge tracking for enhancing their detection.

In \cite{takacs20113d} they propose a 3d mobile augmented reality solution for urban scenes. The 3D MAR system has two main components: offline database processing, and online image matching, tracking and augmentation. One of the steps presented is the contour extraction that is built upon morphological edge detection by subtracting a 1-pixel erosion.

In \cite{reitmayr2006going} the system combines several well-known approaches to provide a robust experience that surpasses each of the individual components alone: an edge-based tracker for accurate localization and gyroscope measurements to deal with fast motions.

In \cite{karlekar2010positioning} a novel approach for user positioning, robust tracking and online 3D mapping for outdoor augmented reality applications. Robust visual tracking is maintained by fusing frame-to-frame and model-to-frame feature matches. Frame-to-frame tracking is accomplished with corner matching while edges are used for model-to-frame registration.

In \cite{blanco2019augmented} they present a methodology to develop immersive AR applications based on the recognition of outdoor buildings. To demonstrate this methodology, a case study focused on the Parliament Buildings National Historic Site in Ottawa, Canada has been conducted.

We believe that using the proposed new operators will improve the quality and number of edges found in each frame by the AR system. This might seem like a small element in the processing chain but this will make the edge tracking better. Improving the edge tracking block will automatically improve the overall performance of an AR application.

\section{Preliminaries}
\label{Sec:preliminaries}

In this section we present the filters we would like to compare. The 3x3 Gx kernels for Sobel \cite{sobel19683x3} , Prewitt \cite{prewitt1970object} and Scharr \cite{scharr2000optimal} are presented in Figure \ref{figure:3_kernel_masks}.

\begin{figure}[h]
\centering
\small{
\centering
\begin{tabular}{cccc}
    $\begin{bmatrix}
        -1 & 0 & 1 \\
        -2 & 0 & 2 \\
        -1 & 0 & 1 \\
    \end{bmatrix}$
    &
    $\begin{bmatrix}
        -1 & 0 & 1 \\
        -1 & 0 & 1 \\
        -1 & 0 & 1 \\
    \end{bmatrix}$
    &
    $\begin{bmatrix}
         -3 & 0 &  3 \\
        -10 & 0 & 10 \\
         -3 & 0 &  3 \\
    \end{bmatrix}$

    \\
    ~
    \\
        Sobel 
    &         
        Prewitt
    &
        Scharr
    \\ 
\end{tabular}
}
\caption{First Order derivative edge operators kernels}
\label{figure:3_kernel_masks}
\end{figure}

The 5x5 Gx kernels for Sobel, Prewitt and Scharr \cite{levkine2012prewitt} are presented in Figure \ref{figure:k5_kernel_masks}.

\begin{figure}[h]
\centering
\small{
\begin{tabular}{ccc}
        $\begin{bmatrix}
             -5 &  -4 & 0 &  4 & 5  \\
             -8 & -10 & 0 & 10 & 8  \\
            -10 & -20 & 0 & 20 & 10 \\
             -8 & -10 & 0 & 10 & 8  \\
             -5 &  -4 & 0 &  4 & 5  \\
        \end{bmatrix}$
    &
        $\begin{bmatrix}
            -2 & -1 & 0 & 1 & 2  \\
            -2 & -1 & 0 & 1 & 2  \\
            -2 & -1 & 0 & 1 & 2  \\
            -2 & -1 & 0 & 1 & 2  \\
            -2 & -1 & 0 & 1 & 2  \\
        \end{bmatrix}$
    &
        $\begin{bmatrix}
            -1 & -1 & 0 & 1 & 1  \\
            -2 & -2 & 0 & 1 & 2  \\
            -3 & -6 & 0 & 6 & 3  \\
            -2 & -2 & 0 & 2 & 2  \\
            -1 & -1 & 0 & 1 & 1  \\
        \end{bmatrix}$
    \\
        ~
    \\
        Sobel Extended
    &
        Prewitt Extended
    &
        Scharr Extended

\end{tabular}
}
\caption{$5x5$ kernels masks}
\label{figure:k5_kernel_masks}
\end{figure}

To determine the exact impact of the filters we use a modified Canny edge detection algorithm \cite{canny1986computational, fang2009study}.
\begin{itemize}
    \item Step 1 Applying Gaussian Filter on the grey-scale image.
    \item Step 2 Apply Otsu transformation on picture to find threshold value
    \item Step 3 Finding the magnitude and orientation of the gradient.
    \item Step 4 Non-maximum suppression.
    \item Step 5 Edge tracking by hysteresis using double threshold. Thresholds are found using Step 2 and applying Formula \ref{formula:th} and \ref{formula:tl}.
\end{itemize}

\begin{align}\label{formula:sigma}
\sigma = 0.33
\end{align}

\begin{align}\label{formula:th}
&\mathit{T}_{h} = max(0, (1.0-\sigma))\times val_{otsu}
\end{align}

\begin{align}\label{formula:tl}
&\mathit{T}_{l} = min(255, (1.0 + \sigma))\times val_{otsu} 
\end{align}

\section{Proposed Filter}
\label{Sec:proposed}

We consider that for the use case we are dealing with, extraction of building features, we must try to detect longer edges for a better reconstruction of the contour and with more attention given to the edges that are vertical and horizontal than on the diagonals. This means that the filter that we propose is constructed based on the distance between pixels and with a bigger weight on the $0^{\circ}$ and $90^{\circ}$ direction of the gradients.

For the first proposed filter we consider the following scheme for creating the weight matrix, that is based on the inverse distance \cite{fotheringham2008sage}, see Formula \ref{formula:scheme} and \ref{formula:scheme_matrix}.

\begin{align}\label{formula:scheme}
&\mathit{w}_{ij} = 
\begin{cases} 
\frac{1}{d_{ij}^{2}},\ $if$\ i \ne j \\
0, \ if\ i=j
\end{cases}
\end{align}

\begin{align}\label{formula:scheme_matrix}
\begin{bmatrix}
\frac{1}{2} &\frac{1}{1} &\frac{1}{2}   \\
\frac{1}{1} &w           &\frac{1}{1}   \\
\frac{1}{2} &\frac{1}{1} &\frac{1}{2}   \\
\end{bmatrix} 
=>
\begin{bmatrix}
1 &2 &1   \\
2 &w &2   \\
1 &2 &1   \\
\end{bmatrix} 
\end{align}

We know that we can use linear approximation to estimate the Formula \ref{formula:estimation} \cite{haralick1981facet}.

\begin{align}\label{formula:estimation}
&\mathit{J}(\alpha, \beta, \gamma) = \sum_{r,c=-(2L+1)}^{2L+1} w_{rc}^2(\alpha * r + \beta * c + \gamma - I(r,c))^2
\end{align}

Using the Local Polynomial Approximation for every pixel we can calculate the kernels equivalents for the weight matrix \cite{levkine2012prewitt, haralick1981facet}. We can see the final kernels in Figure \ref{formula:gx} and Figure \ref{formula:gy_formula}.

\begin{align}\label{formula:gx_formula}
G_{x}(r,c) = \frac{\partial I(r,c)}{\partial r} = \alpha
\end{align}

\begin{align}\label{formula:gy_formula}
G_{y}(r,c) = \frac{\partial I(r,c)}{\partial c} = \beta
\end{align}

\begin{align}\label{formula:gx}
G_{x} = \begin{bmatrix}
-1 &0 &1   \\
-4 &0 &4   \\
-1 &0 &1   \\
\end{bmatrix} 
\end{align}

\begin{align}\label{formula:gy}
G_{y} = \begin{bmatrix}
-1 &-4 &-1   \\
 0 &0 &0   \\
 1 &4 &1   \\
\end{bmatrix} 
\end{align}

Following the scheme presented in Formula \ref{formula:scheme} the 5x5 weight matrix is expanded to result in Formula \ref{formula:scheme_5x5_matrix}. The resulting kernels we present in Figure \ref{formula:gx_5_v1} and Figure \ref{formula:gy_5_v1}.

\begin{align}\label{formula:scheme_5x5_matrix}
\begin{bmatrix}
\frac{1}{8} &\frac{1}{5} &\frac{1}{4}   &\frac{1}{5}  &\frac{1}{8} \\
\frac{1}{5} &\frac{1}{2} &\frac{1}{1}   &\frac{1}{2}  &\frac{1}{5} \\
\frac{1}{4} &\frac{1}{1} &w             &\frac{1}{1}  &\frac{1}{4} \\
\frac{1}{5} &\frac{1}{2} &\frac{1}{1}   &\frac{1}{2}  &\frac{1}{5} \\
\frac{1}{8} &\frac{1}{5} &\frac{1}{4}   &\frac{1}{5}  &\frac{1}{8} \\
\end{bmatrix} 
=>
\begin{bmatrix}
5   &8   &10    &8   &5   \\
8   &20  &40    &20  &8   \\
10  &40  &W     &40  &10   \\
8   &20  &40    &20  &8   \\
5   &8   &10    &8   &5   \\
\end{bmatrix} 
\end{align}

\begin{align}\label{formula:gx_5_v1}
G_{x} = \begin{bmatrix}
-25  &-4  &0  &4   &25   \\
-64  &-10 &0  &10  &64   \\
-100 &-20 &0  &20  &100   \\
-64  &-10 &0  &10  &64   \\
-25  &-4  &0  &4   &25   \\
\end{bmatrix} 
\end{align}

\begin{align}\label{formula:gy_5_v1}
G_{y} = \begin{bmatrix}
-25  &-4  &-100  &-4   &-25   \\
-64  &-10 &-20   &-10  &-64   \\
  0  &0   &0     &0    &0   \\
 64  &10  &20    &10   &64   \\
 25  &4  &100  &4   &25   \\
\end{bmatrix} 
\end{align}

We propose the second filter, based on the radial distance weights \cite{fotheringham2008sage} presented in Formula \ref{formula:scheme_v2}. Following the same logic presented above we obtain the Formula \ref{formula:scheme_matrix_v2} that represents the 3x3 weight matrix, Formula \ref{formula:gx_v2} the Gx and Formula \ref{formula:gy_v2} the Gy kernels.

\begin{align}\label{formula:scheme_v2}
&\mathit{w}_{ij} = 
\begin{cases} 
d_{max},\ $if$\ i=0\ or\  j=0 \\
\frac{d_{max}}{2},\ $if$\ i\ne0\ or\  j\ne0 \\
0, \ if\ i=j
\end{cases}
\end{align}

\begin{align}\label{formula:scheme_matrix_v2}
\begin{bmatrix}
\frac{1}{2} &\frac{1}{1} &\frac{1}{2}   \\
\frac{1}{1} &w           &\frac{1}{1}   \\
\frac{1}{2} &\frac{1}{1} &\frac{1}{2}   \\
\end{bmatrix} 
=>
\begin{bmatrix}
1 &2 &1   \\
2 &W &2   \\
1 &2 &1   \\
\end{bmatrix} 
\end{align}

\begin{align}\label{formula:gx_v2}
G_{x} = \begin{bmatrix}
-1 &0 &1   \\
-4 &0 &4   \\
-1 &0 &1   \\
\end{bmatrix} 
\end{align}

\begin{align}\label{formula:gy_v2}
G_{y} = \begin{bmatrix}
-1 &-4 &-1   \\
 0 &0 &0   \\
 1 &4 &1   \\
\end{bmatrix} 
\end{align}

In Formula \ref{formula:scheme_5x5_matrix_v2} we present the scheme for the $5x5$ kernel expansion. In Formula \ref{formula:gx_5_v2} the Gx kernel and in Formula \ref{formula:gy_5_v2} the Gy kernel.

\begin{align}\label{formula:scheme_5x5_matrix_v2}
\begin{bmatrix}
2 &2 &4   &2  &2 \\
2 &2 &4   &2  &2 \\
4 &4 &w   &4  &4 \\
2 &2 &4   &2  &2 \\
2 &2 &4   &2  &2 \\
\end{bmatrix} 
=>
\begin{bmatrix}
1   &1   &2    &1   &1   \\
1   &1   &2    &1   &1   \\
2   &2   &W    &2   &2   \\
1   &1   &2    &1   &1   \\
1   &1   &2    &1   &1   \\
\end{bmatrix} 
\end{align}

\begin{align}\label{formula:gx_5_v2}
G_{x} = \begin{bmatrix}
-2  &-1  &0  &1   &2   \\
-2  &-1  &0  &1   &2   \\
-8 &-4 &0  &4  &8   \\
-2  &-1  &0  &1   &2   \\
-2  &-1  &0  &1   &2   \\
\end{bmatrix} 
\end{align}

\begin{align}\label{formula:gy_5_v2}
G_{y} = \begin{bmatrix}
-2  &-2  &-8  &-2   &-2   \\
-1  &-1  &-4  &-1   &-1   \\
  0  &0   &0     &0    &0   \\
1  &1  &4  &1   &1   \\
2  &2  &8  &2   &2   \\
\end{bmatrix} 
\end{align}

\section{Simulation results}
\label{Sec:sim}

First, we will compare the result we obtain by using the standard formula for calculating the gradient magnitude presented in Formula \ref{formula:mag}.

\begin{align}\label{formula:mag}
    G[f(x,y)] = \sqrt{G_x^2 + G_y^2} \approx |G_x| + |G_y|
\end{align}

\begin{figure}[h]
    \centering
    \begin{minipage}{0.33\textwidth}
        \centering
        \includegraphics[width=\textwidth]{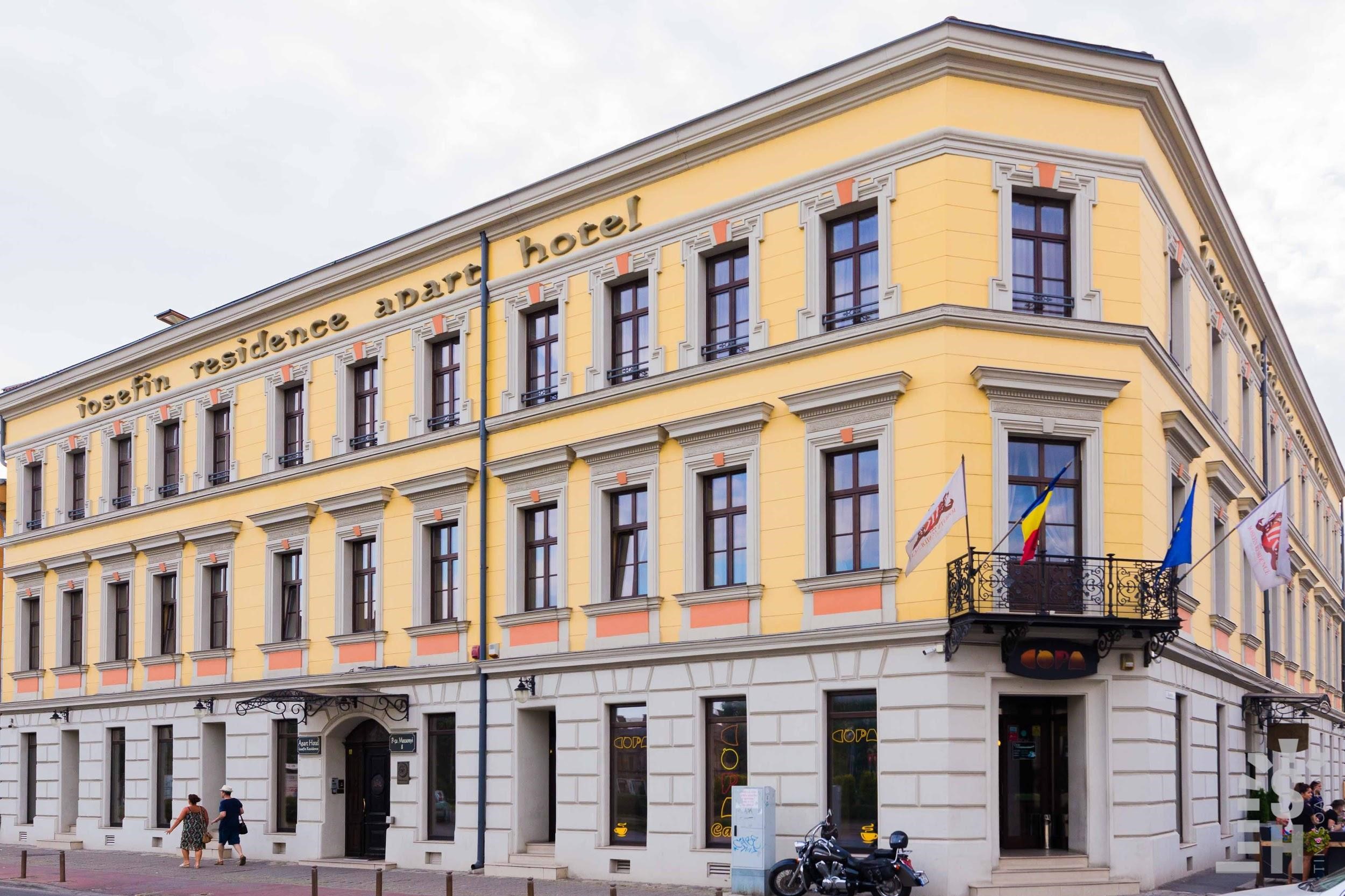} 
        \caption{Original image}
        \label{fig:fig1}
    \end{minipage}\hfill
    \begin{minipage}{0.33\textwidth}
        \centering
        \includegraphics[width=\textwidth]{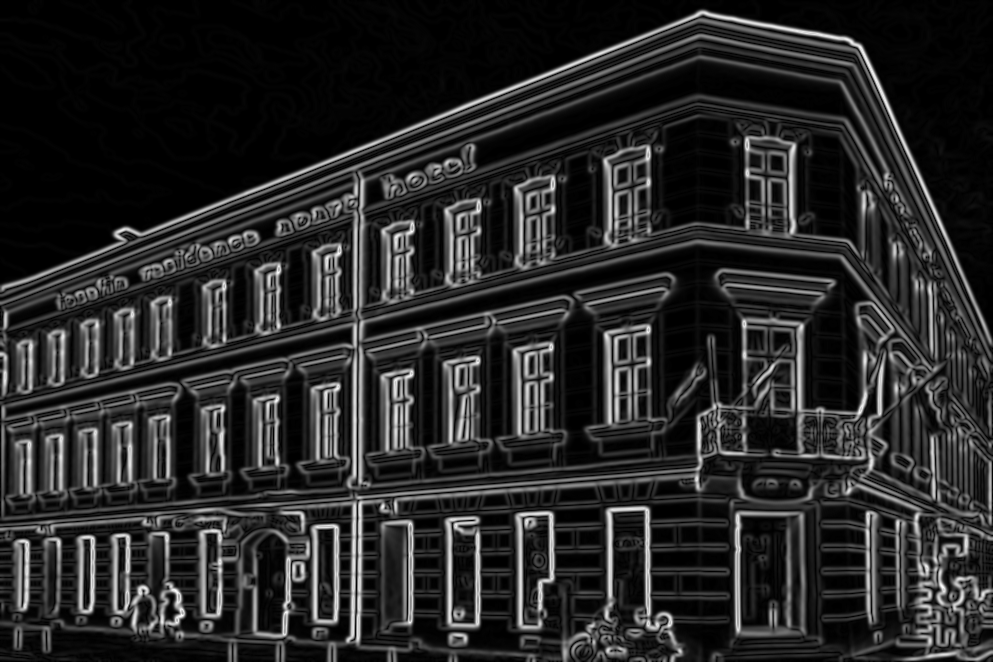} 
        \caption{Proposed 3x3}
        \label{fig:fig2}
    \end{minipage}\hfill
    \begin{minipage}{0.33\textwidth}
        \centering
        \includegraphics[width=\textwidth]{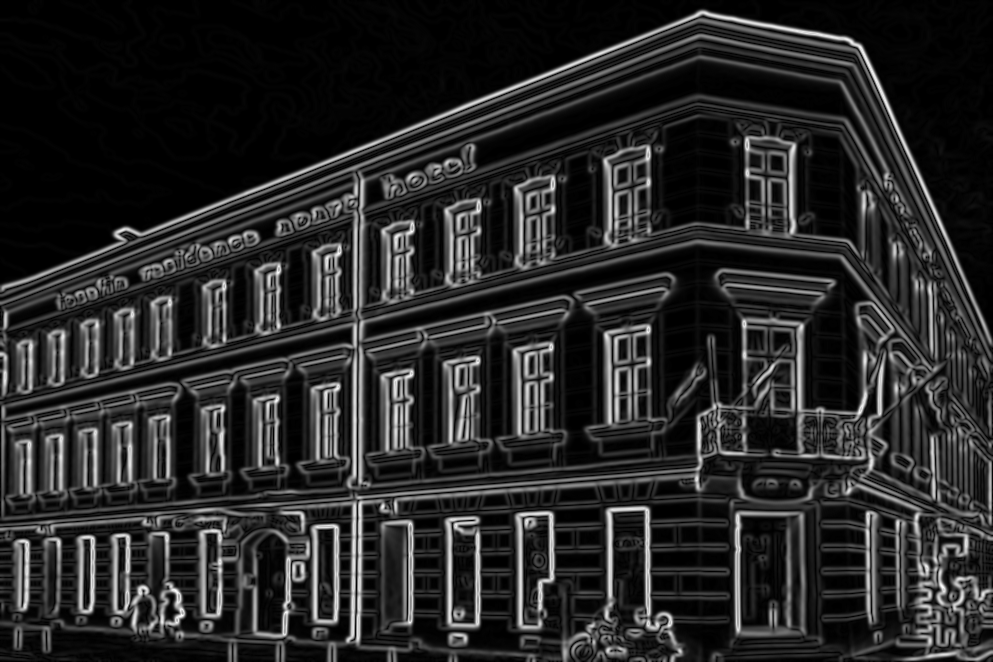} 
        \caption{Proposed A 5x5}
        \label{fig:fig3}
    \end{minipage}\hfill
\end{figure}

\begin{figure}[h]
    \centering
    \begin{minipage}{0.33\textwidth}
        \centering
        \includegraphics[width=\textwidth]{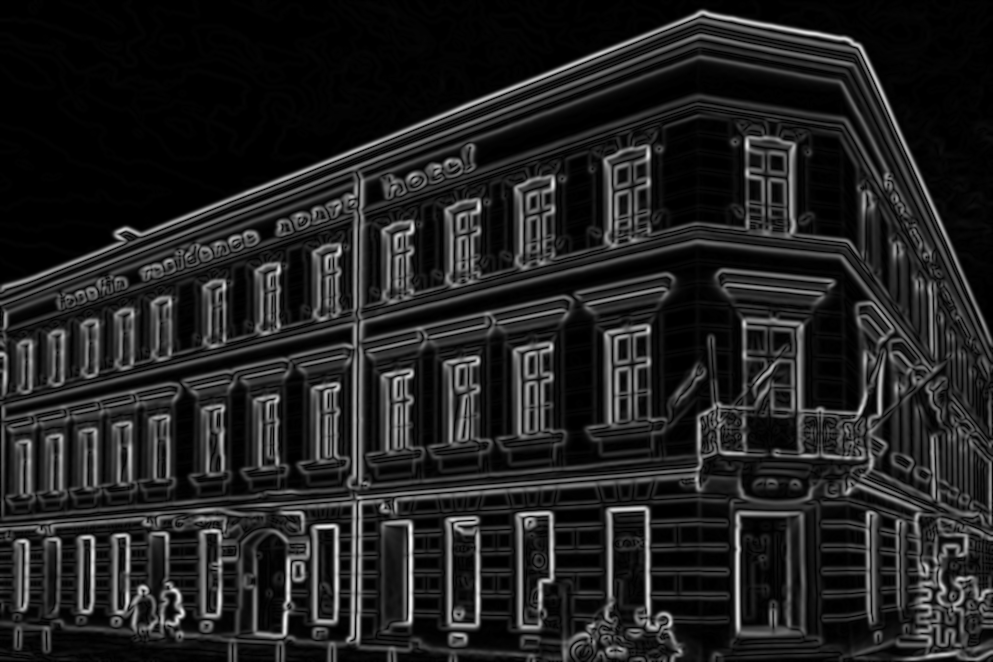} 
        \caption{Proposed B 5x5}
        \label{fig:fig4}
    \end{minipage}\hfill
    \begin{minipage}{0.33\textwidth}
        \centering
        \includegraphics[width=\textwidth]{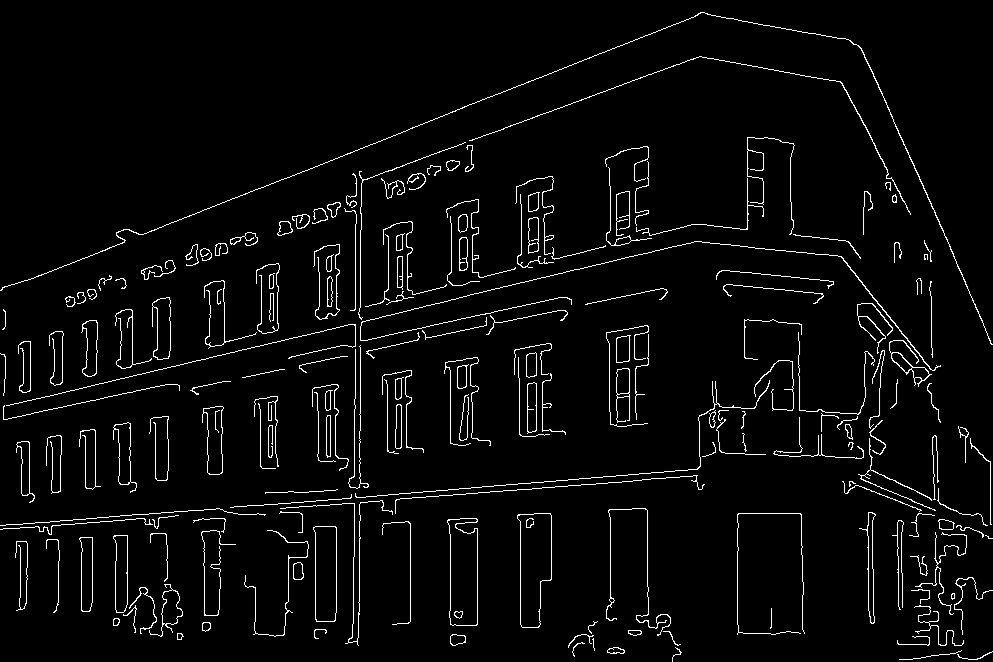} 
        \caption{Canny Sobel 3x3}
        \label{fig:fig5}
    \end{minipage}\hfill
    \begin{minipage}{0.33\textwidth}
        \centering
        \includegraphics[width=\textwidth]{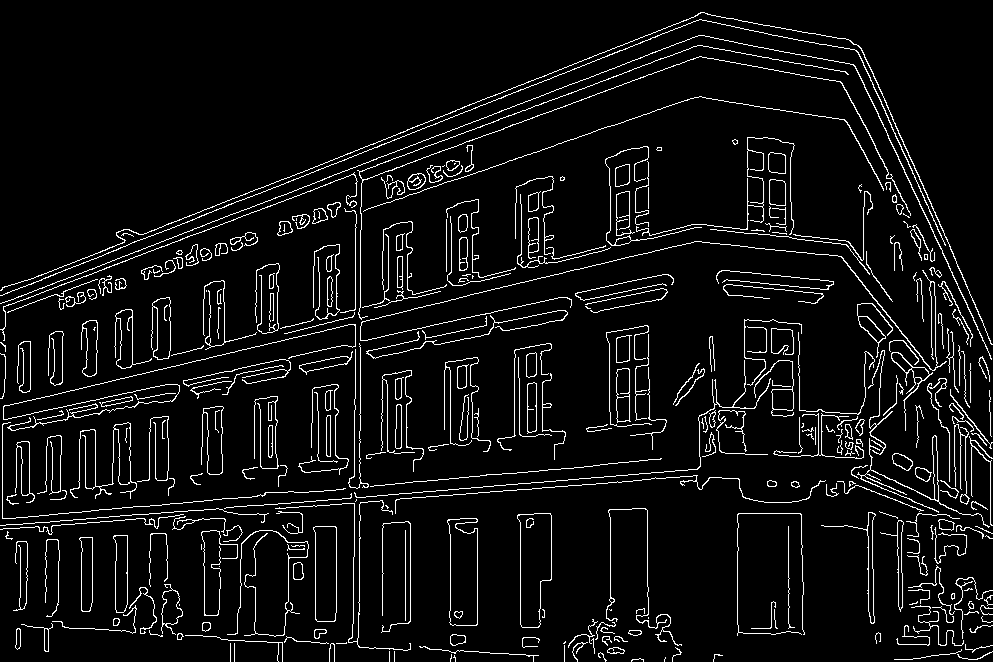} 
        \caption{Canny Proposed A\_B 3x3}
        \label{fig:fig6}
    \end{minipage}\hfill
\end{figure}

We will use the kernels presented in Section \ref{Sec:proposed} and apply Formula \ref{formula:mag} and obtain the gradient magnitude that are presented in Figures \ref{fig:fig2}-\ref{fig:fig4}.

Visually we can determine from Figures \ref{fig:fig2}-\ref{fig:fig4} that the 3x3 filters produce more concentrated edges than the 5x5 filters. But the 5x5 filters tend to lose a certain amount of detail on behalf of edges relevant for contour extraction. 

As stated in section 3 to have a better idea about what edges have added value for a feature extraction based on them we would look over the results we got using the Canny algorithm \cite{canny1986computational, fang2009study}.

\begin{figure}[h]
    \centering
    \begin{minipage}{0.33\textwidth}
        \centering
        \includegraphics[width=\textwidth]{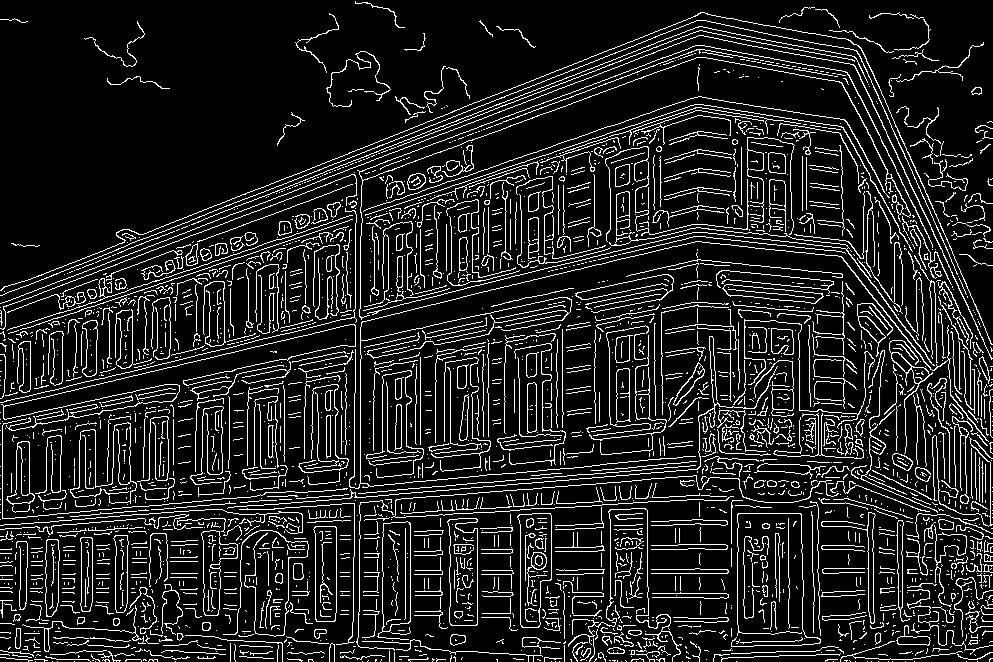} 
        \caption{Canny Sobel 5x5}
        \label{fig:fig7}
    \end{minipage}\hfill
    \begin{minipage}{0.33\textwidth}
        \centering
        \includegraphics[width=\textwidth]{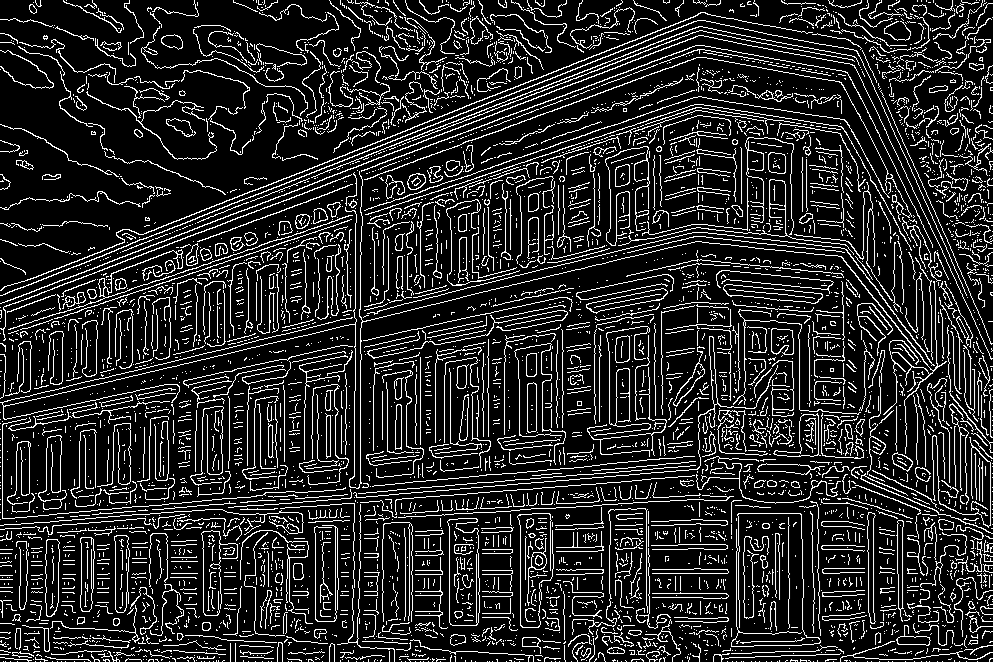} 
        \caption{Canny Proposed A 5x5}
        \label{fig:fig8}
    \end{minipage}\hfill
    \begin{minipage}{0.33\textwidth}
        \centering
        \includegraphics[width=\textwidth]{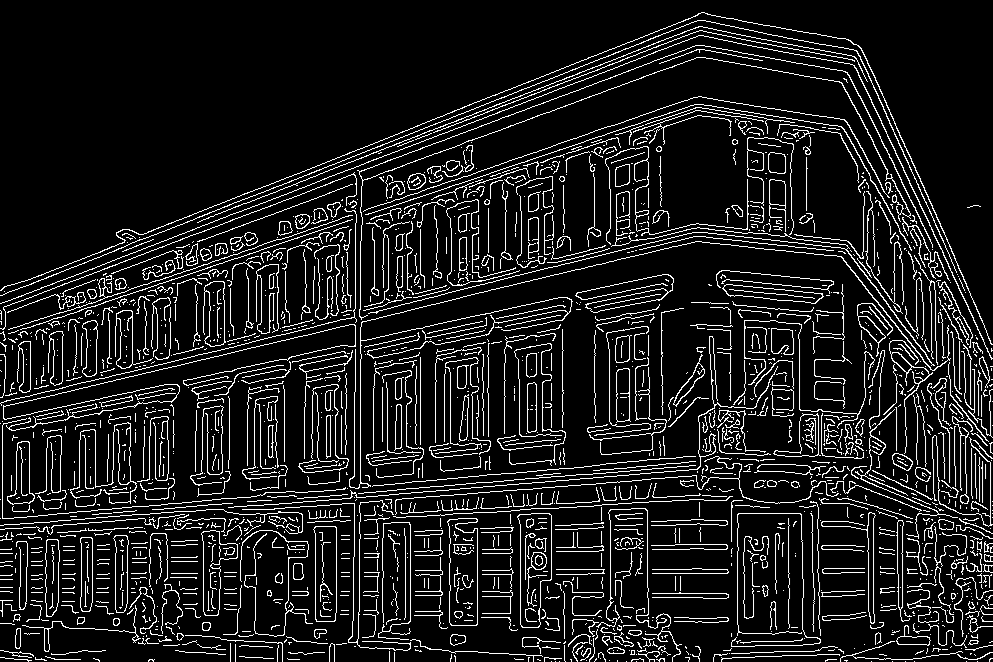} 
        \caption{Canny Proposed B 5x5 }
        \label{fig:fig9}
    \end{minipage}\hfill
\end{figure}

\begin{figure}[h]
    \centering
    \begin{minipage}{0.33\textwidth}
        \centering
        \includegraphics[width=\textwidth]{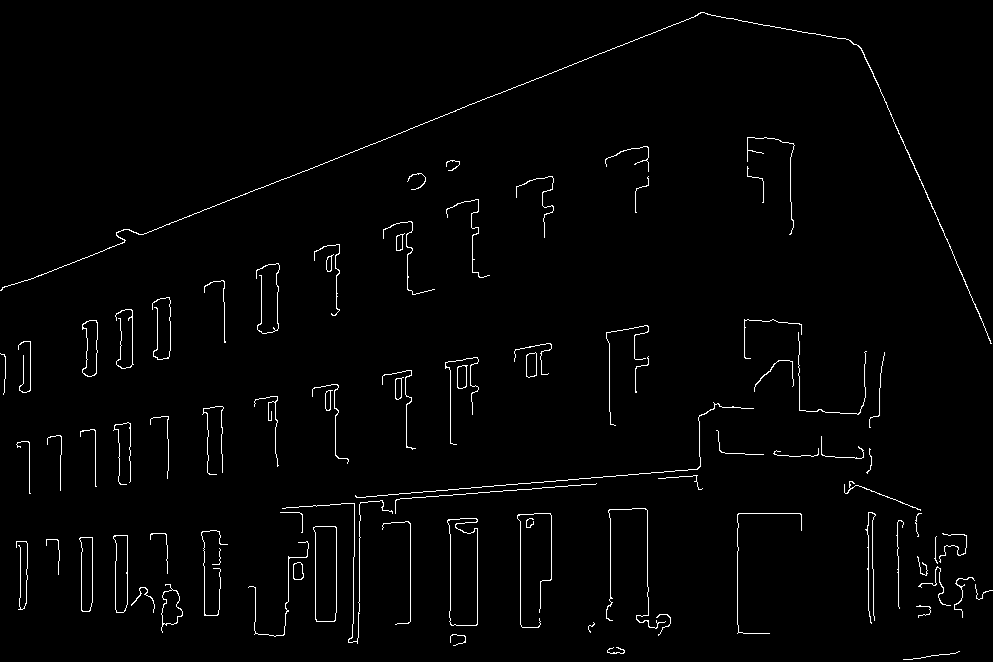} 
        \caption{Canny Prewitt 3x3}
        \label{fig:fig10}
    \end{minipage}\hfill
    \begin{minipage}{0.33\textwidth}
        \centering
        \includegraphics[width=\textwidth]{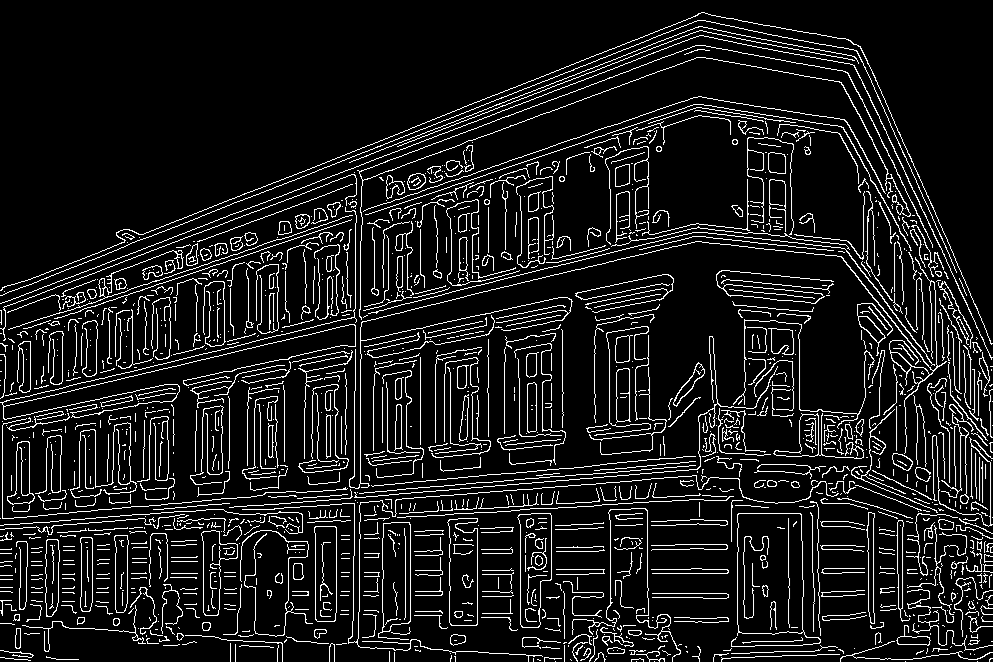} 
        \caption{Canny Prewitt 5x5}
        \label{fig:fig11}
    \end{minipage}\hfill
    \begin{minipage}{0.33\textwidth}
        \centering
        \includegraphics[width=\textwidth]{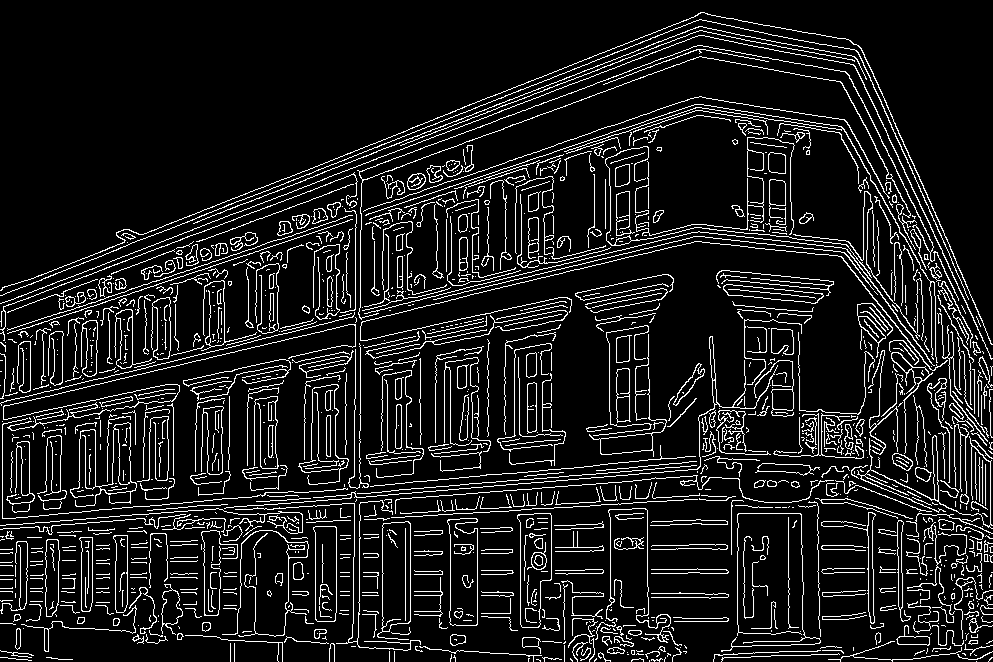} 
        \caption{Canny Scharr 3x3 }
        \label{fig:fig12}
    \end{minipage}\hfill
\end{figure}

\begin{figure}[h]
    \centering
    \begin{minipage}{0.33\textwidth}
        \centering
        \includegraphics[width=\textwidth]{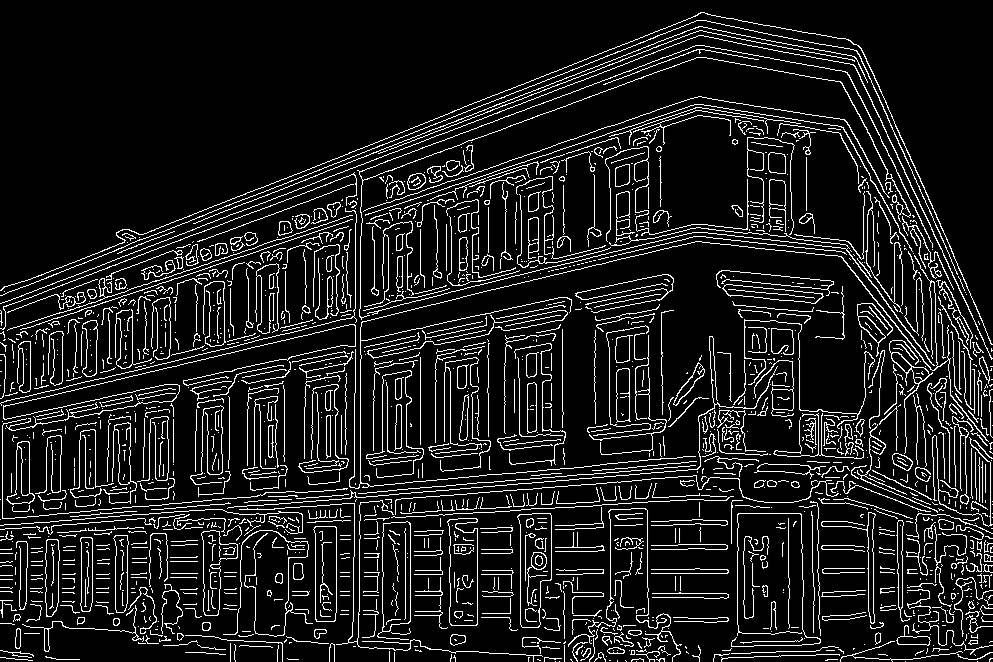} 
        \caption{Canny Scharr 5x5}
        \label{fig:fig13}
    \end{minipage}\hfill
    \begin{minipage}{0.33\textwidth}
        \centering
        \includegraphics[width=\textwidth]{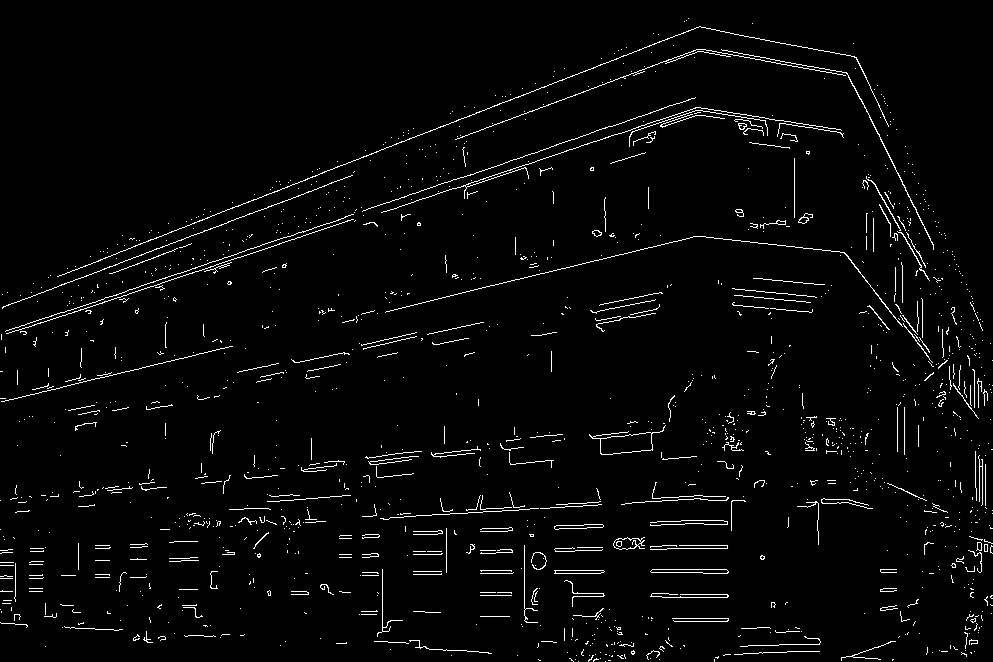} 
        \caption{Canny Proposed A vs Sobel}
        \label{fig:fig14}
    \end{minipage}\hfill
    \begin{minipage}{0.33\textwidth}
        \centering
        \includegraphics[width=\textwidth]{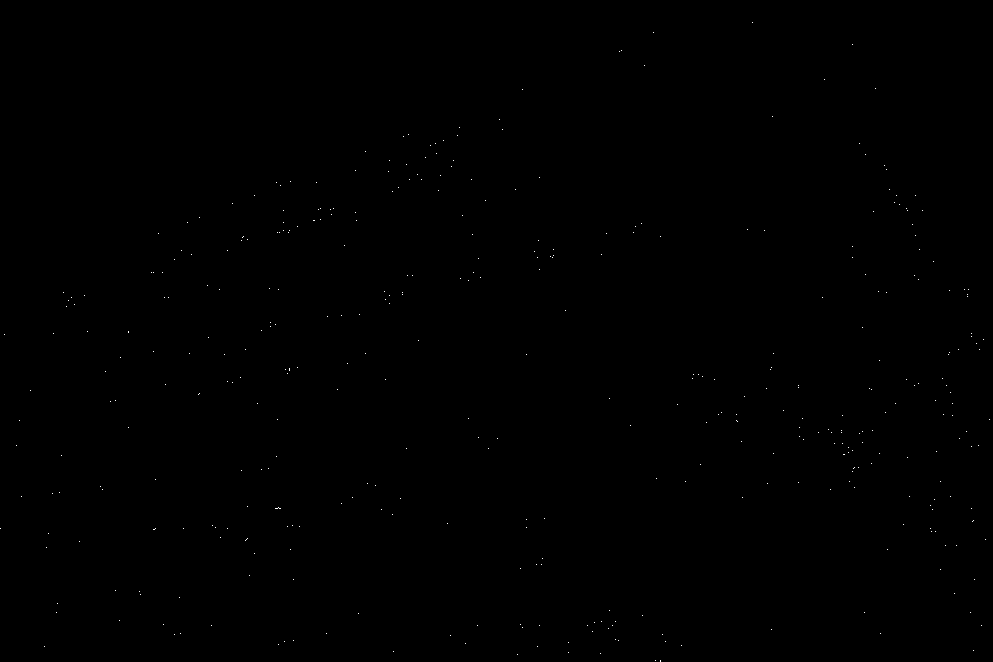} 
        \caption{Canny Sobel vs Proposed A }
        \label{fig:fig15}
    \end{minipage}\hfill
\end{figure}

To obtain results that we can future use for contour detection or facade detection should be a balance between edges detected from the outer boundary of the building and facade details. The edges that do not belong to the buildings should be preferably discarded in this step. As we can see in Figures \ref{fig:fig5}-\ref{fig:fig13} the filters 3x3 obtain better results than the 5x5 ones.

Figure \ref{fig:fig7}, \ref{fig:fig8} show a considerable amount of edges detected that we can consider noise and those edges will strongly affect any future use of those results.

Figure \ref{fig:fig9}, \ref{fig:fig11}, \ref{fig:fig12}, \ref{fig:fig13} show good results in filtering out edges that are not in our interest but we can observe an extra amount of details on the facade detected that is not a benefit in future work.

The proposed filters and Sobel 3x3 \cite{sobel19683x3} used in the Canny algorithm \cite{canny1986computational, fang2009study}, shown in Figure \ref{fig:fig5}, \ref{fig:fig6}, show good results in maintaining a balance between filtering out edges that we can consider noise and edges that we can future use for contour and edges that can be used for facade detection. 

Naturally, we desired to better understand our results, so we took a dataset of 30 images that we used for Spotlight Heritage Timisoara project and compared the results obtained between Canny using Sobel 3x3 and Canny using our proposed filter. 

\begin{figure}[h]
    \centering
    \begin{minipage}{0.49\textwidth}
        \centering
        \includegraphics[width=\textwidth]{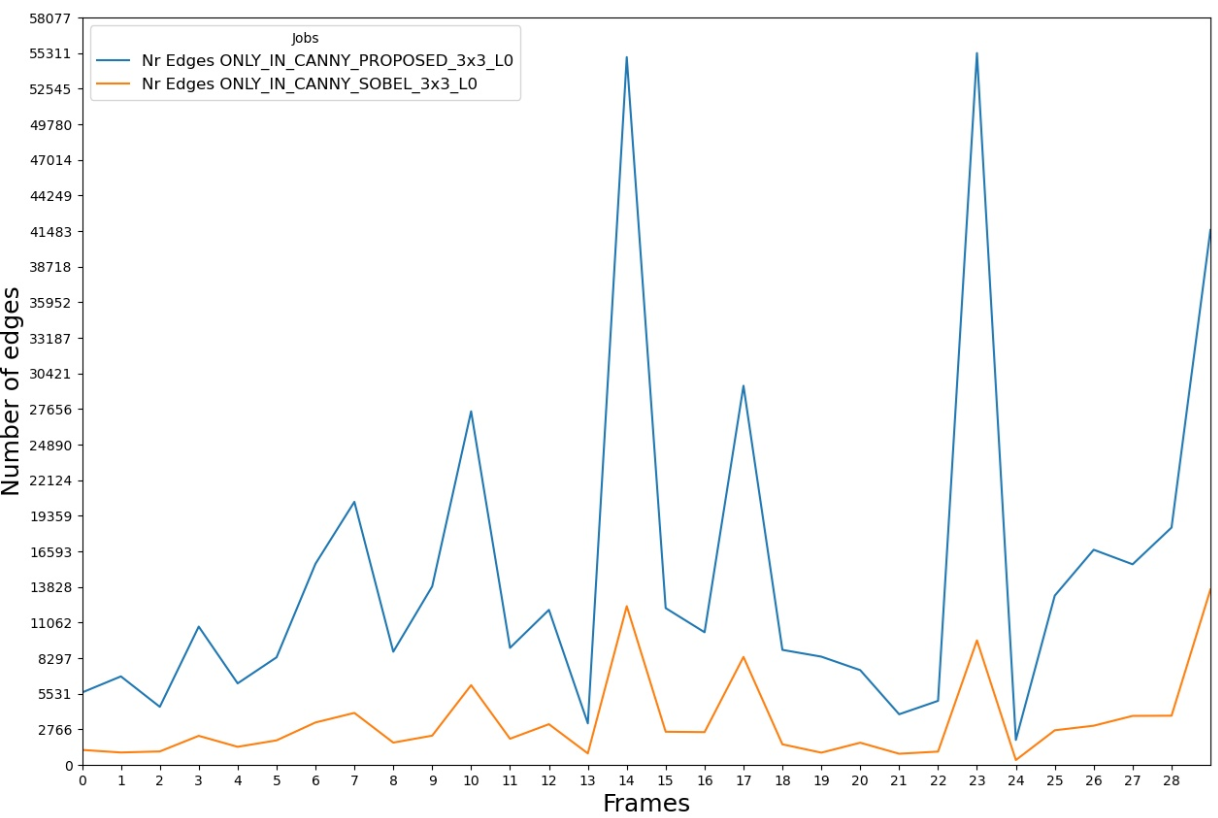} 
        \caption{Number of edges found}
        \label{fig:fig16}
    \end{minipage}\hfill
    \begin{minipage}{0.49\textwidth}
        \centering
        \includegraphics[width=\textwidth]{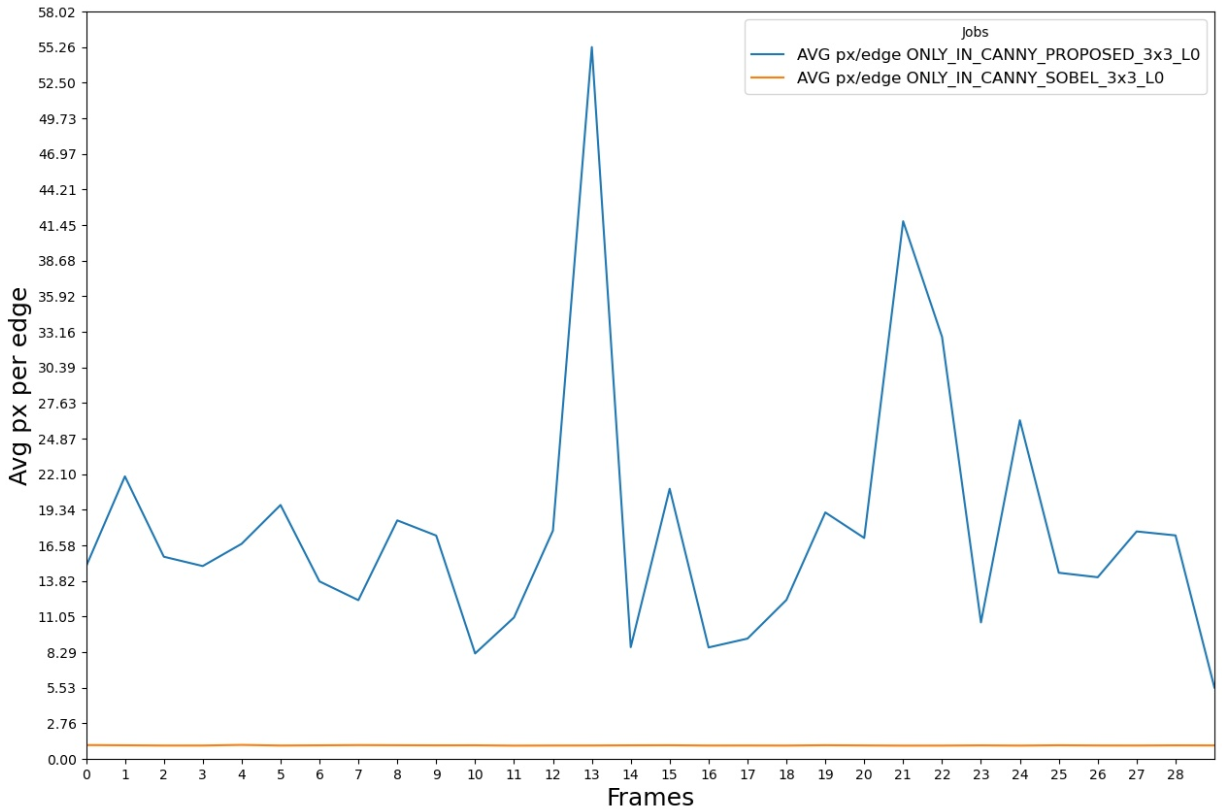} 
        \caption{Average pixels per edge}
        \label{fig:fig17}
    \end{minipage}\hfill
\end{figure}

The first result we saw was the fact that the number of edge pixels was always bigger, a fact that was observed from the images presented in this section. But we wanted to test our statement that the edge pixels we obtain form more and longer edges. To do that we used an eight-neighbor connectivity algorithm \cite{bhattacharya2000convexity} and we can observe in Figure \ref{fig:fig16} and \ref{fig:fig17} the results.

In Figure \ref{fig:fig16} we observe that in every case our Canny using our proposed filter obtains more edge pixels that formed more edges than the Canny using Sobel filter. This fact alone does not prove that the edges found are useful but if we look on Figure \ref{fig:fig17} we can observe that those edges group in longer edges. We can clearly observe that the edges pixels found by the Sobel filter and not by our filter are sporadic pixels.

\section{Conclusion and Future work}
\label{Sec:conclusion}

It is difficult to design a general edge detection algorithm which performs well in many contexts and captures the requirements of subsequent processing stages. With this research paper, we aimed to find a better fitted first derivative operator for building contour and facade edge detection.

We proposed two filters that have the same kernels for 3x3 but with different kernels for 5x5. As we can observe from Figure \ref{fig:fig15} our proposed filter managed to detect the same edges using the classical Sobel filter when used in the Canny algorithm.

As future work we plan to try the 3x3 filter operator, that obtains the best results, on a bigger data set of pictures with an official benchmarking tool for edge detection so we can assert statistically the improvements the proposed filter brings in this use case. At this point we analyzed the differences this filter brought to edge detection feature, as future work we would like to see the effect it has in contour detection algorithm, respective in facade detection algorithm.

This work will be valuable in improving the augmented reality mobile application of the Spotlight Heritage Timisoara project that the authors of this paper are involved in.

\bibliographystyle{ieeetr}



\end{document}